\documentclass{article}
\usepackage[main,final]{neurips_2026}

\usepackage[utf8]{inputenc} 
\usepackage[T1]{fontenc}    
\usepackage{hyperref}       
\usepackage{url}            
\usepackage{booktabs}       
\usepackage{amsfonts}       
\usepackage{nicefrac}       
\usepackage{microtype}      
\usepackage[table]{xcolor}         %
\usepackage{graphicx}
\usepackage{multirow}
\usepackage{wrapfig}
\usepackage{enumitem}

\title{When Latents Forget Pixels: Restoring Fidelity in Diffusion Transformer Super-Resolution}
\newcommand{\yuyao}[1]{\textcolor{black}{#1}}
\newcommand{\sy}[1]{\textcolor{black}{#1}}

\author{
 \textbf{Yu Shi}\\
Dartmouth College\\
\and
\textbf{Yuyao Zhang}\\
Dartmouth College\\
\and
\textbf{Yu-Wing Tai}\\
Dartmouth College\\
}
\begin{document}
\maketitle
\begin{center}\vspace{-0.5in}
\textbf{Project:} \url{https://rain-sy.github.io/projects/PGSR/}
\end{center}

\vspace{-0.1in}
\begin{figure}[h]
    \centering
    \includegraphics[width=.95\linewidth]{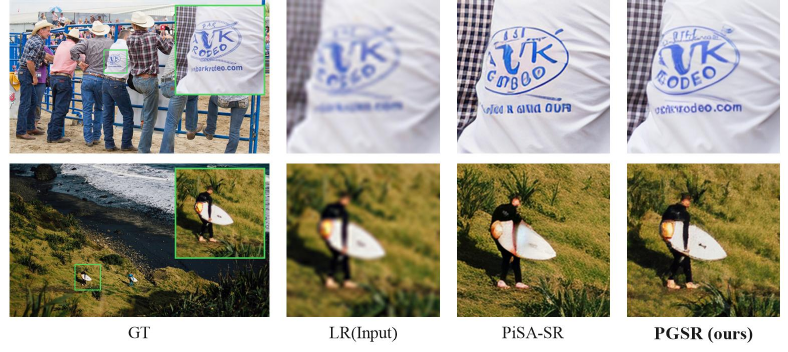}
    \\
    \vspace{-0.1in}
    \caption{
    Super-resolution results using \textbf{PGSR} compared to one of the other State-of-the-art pixel-guided methods.
    While it produces visually plausible textures, it hallucinates structures inconsistent with the input (e.g., incorrect texts or unexpectedly sharpened edges).
    \yuyao{Our method preserves LR pixel features before VAE compression and uses them to guide latent trajectory and ground final rendering, producing sharper results that remain faithful to the LR observation.}
    }
    \label{fig:teaser}
\end{figure}

\begin{abstract}
Image super-resolution (SR) with large generative models has recently achieved remarkable perceptual quality, yet maintaining fidelity to the LR observation remains challenging. In particular, we observe that diffusion transformers (DiTs) built on latent representations suffer from a critical limitation: the compression bottleneck of the VAE weakens fine-grained spatial information, leading to hallucinated details that are weakly grounded in the input image. In this work, we revisit generative SR from a representation perspective and propose a \textbf{pixel-grounded super-resolution (PGSR)} framework that preserves LR-observed pixel evidence before VAE compression and reuses it throughout restoration. Instead of relying solely on the compressed latent condition, PGSR extracts pre-VAE pixel evidence from the upsampled LR image and reuses it at two stages. First, \textbf{Condition-Side Trajectory Guidance} fuses LR-derived pixel evidence with the latent LR condition to guide the latent restoration trajectory. Second, \textbf{Decoder-Side Pixel Grounding} injects multi-scale pixel features into the frozen VAE decoder to ground the final rendering with LR-observed cues. To efficiently adapt large pretrained DiT models, we keep the latent autoencoder and main flow-matching backbone frozen, and train only lightweight restoration modules. We further study an efficient local-window attention variant for improved high-resolution efficiency and scalability. Extensive experiments demonstrate that PGSR improves the realism--fidelity trade-off and produces more faithful, visually convincing results than existing latent generative SR approaches. Code is available at \url{https://github.com/Rain-sy/PGSR}.
\end{abstract}

\section{Introduction}

Image super-resolution (SR) aims to recover high-frequency details from a low-resolution (LR) input while preserving fidelity to the observed image. Recent approaches leverage large pretrained generative models, including diffusion and transformer-based architectures, to synthesize realistic textures that exceed the capabilities of regression- or GAN-based methods~\citep{ledig2017srgan, wang2018esrgan, saharia2021sr3, wang2023stablesr, wu2023seesr}. However, despite their strong perceptual quality, these methods still struggle to remain consistent with the input, often hallucinating details that are only weakly supported by the LR image.

In this work, we identify a key but underexplored reason for this failure: \emph{latent generative models operate on compressed representations that can weaken fidelity-critical pixel evidence}. Modern diffusion transformers and latent diffusion models rely on a VAE tokenizer~\citep{kingma2014autoencoding, rombach2022ldm} to map images into a compact latent space. While this enables efficient high-resolution generation, it also introduces an information bottleneck that removes subtle spatial cues and high-frequency structures essential for faithful SR. As illustrated in Fig.~\ref{fig:teaser}, even when conditioned on the LR input, latent-only generation tends to produce visually convincing but incorrect details, revealing a fundamental mismatch between latent generative priors and pixel-level restoration objectives.

This perspective suggests that improving SR is not merely a matter of stronger conditioning, but of \emph{\yuyao{keeping LR-observed image evidence accessible before it is compressed into the latent space.}} Prior works have attempted to mitigate this issue by introducing pixel-aware guidance or additional control mechanisms within diffusion-based SR pipelines~\citep{wu2023seesr, yang2023pasd, sun2025pisasr, arora2025guidesr}. While effective, most of these designs still use pixel evidence primarily as an external condition for the latent denoiser. They therefore leave two fidelity gaps only partially addressed: the latent guided trajectory may drift away from LR-observed structure, and the frozen decoder may further lose spatial detail during final rendering.

To this end, we propose a \textbf{pixel-grounded super-resolution (PGSR)} framework that explicitly \yuyao{preserves LR-observed pixel cues before VAE compression and reuses them throughout restoration}. We refer to this design as Pixel-Grounded Guidance: pixel evidence denotes image-domain features extracted from the LR input before VAE compression and used to restore fidelity-critical spatial information. PGSR realizes this idea through \yuyao{two complementary mechanisms}. First, Condition-Side Trajectory Guidance fuses LR-derived pixel evidence into the ControlNet condition to anchor the flow trajectory. Second, Decoder-Side Pixel Grounding injects multi-scale pixel features into the frozen VAE decoder to recover fine spatial structures at rendering time. \yuyao{In this way, the pretrained latent backbone provides strong semantic and generative priors, while the pixel-grounded pathway constrains restoration with LR-observed image evidence.}

Importantly, our design differs from existing pixel-guided SR methods in that pixel evidence is not only appended as an auxiliary condition. Instead, it is used to ground both what content is generated in latent space and how that content is decoded back into pixels. By explicitly modeling the residual information missing from the latent space, our approach directly targets the root cause of fidelity degradation in generative SR.

To enable practical deployment on large pretrained models, we adopt a parameter-efficient adaptation strategy based on a ControlNet-style architecture~\citep{zhang2023controlnet}, where the diffusion transformer backbone remains frozen and only lightweight modules are trained. We further adapt the frozen flow-matching transformer with low-rank adapters and local-window attention to improve high-resolution efficiency.

In summary, our contributions are three-fold:
\begin{itemize}[leftmargin=20pt, itemsep=2pt, topsep=2pt, parsep=2pt, partopsep=0pt]
    \item We identify latent compression as a fundamental bottleneck in generative super-resolution, and reinterpret SR as a mismatch between latent generative priors and pixel-faithful restoration.
    \item We propose Pixel-Grounded Guidance, a dual-stage design that reuses pre-VAE pixel evidence to guide both latent trajectory formation and final image decoding, improving fidelity throughout the restoration process.
    \item We present a practical and efficient implementation on pretrained latent flow-matching models, using lightweight adaptation and local attention to achieve a better realism--fidelity trade-off.
\end{itemize}

\section{Related Work}

\noindent\textbf{Generative Super-Resolution and the Realism--Fidelity Trade-off.} Early super-resolution (SR) methods~\cite{chang04sr,tai06sr,sun08sr,tai10sr} focused on distortion minimization, optimizing metrics such as PSNR and SSIM~\citep{wang2004ssim}, with representative models including SRCNN~\citep{dong2014srcnn}, VDSR~\citep{kim2016accurate}, and transformer-based restorers such as SwinIR~\citep{liang2021swinir}. While effective on distortion metrics, these approaches often produce over-smoothed results.

Perceptual SR methods emphasize visual realism instead. SRGAN~\citep{ledig2017srgan} and ESRGAN~\citep{wang2018esrgan} show that adversarial training and perceptual losses~\citep{johnson2016perceptual, zhang2018lpips} yield sharper, more realistic textures. This shift is formalized by the perception--distortion trade-off~\citep{blau2018perception}, where improved perceptual quality often reduces fidelity. GLEAN~\citep{chan2021glean} further leverages pretrained generative priors for large-scale SR.

For real-world SR, degradation modeling is critical. BSRGAN and Real-ESRGAN~\citep{zhang2021bsrgan, wang2021realesrgan} highlight the importance of realistic blur--noise--compression pipelines. Diffusion-based SR improves perceptual quality and stability: SR3~\citep{saharia2021sr3} and DDRM~\citep{kawar2022ddrm} introduce diffusion priors, while ResShift~\citep{yue2023resshift}, SinSR~\citep{wang2023sinsr}, and OSEDiff~\citep{wu2024osediff} improve efficiency. Recent work explores rectified-flow formulations such as FlowSR and FluxSR~\citep{xu2025flowsr, li2025fluxsr}. Another direction adapts pretrained generative priors, such as StableSR~\citep{wang2023stablesr}, DiffBIR~\citep{lin2024diffbir}, XPSR~\citep{xpsr2024}, and SeeSR~\citep{wu2023seesr}, to enhance restoration via semantic and structural conditioning.

Despite these advances, the realism--fidelity trade-off persists: stronger generative priors improve perceptual quality by relaxing constraints from the LR input, often introducing hallucinated details.

\noindent\textbf{Limitations of Latent-Space Diffusion for Restoration.} Latent diffusion models (LDMs)~\citep{rombach2022ldm} enable scalable high-resolution generation by operating on compressed representations, but this efficiency introduces an information bottleneck that discards fine-grained spatial details critical for faithful restoration. Recent analyses highlight a reconstruction--generation trade-off, where stronger compression benefits synthesis but degrades input fidelity~\citep{yao2025reconstruction}.

This issue is especially pronounced in SR, where subtle pixel-level structures must be preserved. Prior works mitigate it by strengthening conditioning or modifying inference~\citep{wang2023stablesr, lin2024diffbir, wang2023sinsr}, yet the underlying loss of pixel-domain information remains unresolved. Recent studies therefore revisit pixel-space modeling. Simple Diffusion~\citep{hoogeboom2023simple} and Hourglass Diffusion Transformers~\citep{crowson2024hourglass} show that pixel-space generation can scale with proper design, while JiT~\citep{li2025jit} and Latent Forcing~\citep{baade2026latentforcing} further question the need for aggressive compression.

These findings suggest that preserving or reintroducing pixel-space information is key to faithful reconstruction. Rather than discarding latent priors, our approach compensates for their information loss by reusing pre-VAE pixel evidence throughout restoration.

\noindent\textbf{Pixel-Guided and Grounded Super-Resolution.} To mitigate fidelity loss, recent SR methods incorporate stronger conditioning into pretrained generative models. ControlNet~\citep{zhang2023controlnet} enables spatially aligned conditioning and is widely used in restoration pipelines such as DiffBIR~\citep{lin2024diffbir}. Adapter-based methods, including T2I-Adapter~\citep{mou2024t2iadapter} and LoRA~\citep{hu2021lora}, provide parameter-efficient task-specific guidance.

More explicitly, pixel-aware SR methods aim to preserve local structures. PASD~\citep{yang2023pasd} introduces pixel-level guidance into latent diffusion. SeeSR~\citep{wu2023seesr}, FaithDiff~\citep{chen2025faithdiff}, and GenDR~\citep{wang2025gendr} improve fidelity via semantic and structural conditioning. PiSA-SR~\citep{sun2025pisasr} disentangles pixel fidelity and semantic enhancement with dual LoRA modules, while GuideSR~\citep{arora2025guidesr} adds a full-resolution guidance branch. However, these methods typically treat pixel information as auxiliary to latent generation, leaving two issues: (1) latent trajectories may drift from LR structure, and (2) VAE decoding can further lose spatial detail.

Our proposed Pixel-Grounded Super-Resolution (PGSR) addresses both issues by reusing pre-VAE pixel evidence at two stages: guiding the latent restoration trajectory and grounding the final VAE decoding. This dual design directly targets the information bottleneck of latent generative SR, enabling more faithful and consistent reconstruction.

\section{Methodology}

In this section, we present PGSR as a pixel-grounded latent flow-matching framework for super-resolution. Instead of using the LR image only after VAE encoding, PGSR leverages pre-VAE pixel evidence to guide both latent trajectory formation and final rendering, addressing two fidelity gaps: VAE compression weakens LR structure, and the frozen decoder may lose spatial detail. 

\begin{figure*}[t]
    \centering
    \includegraphics[width=\textwidth]{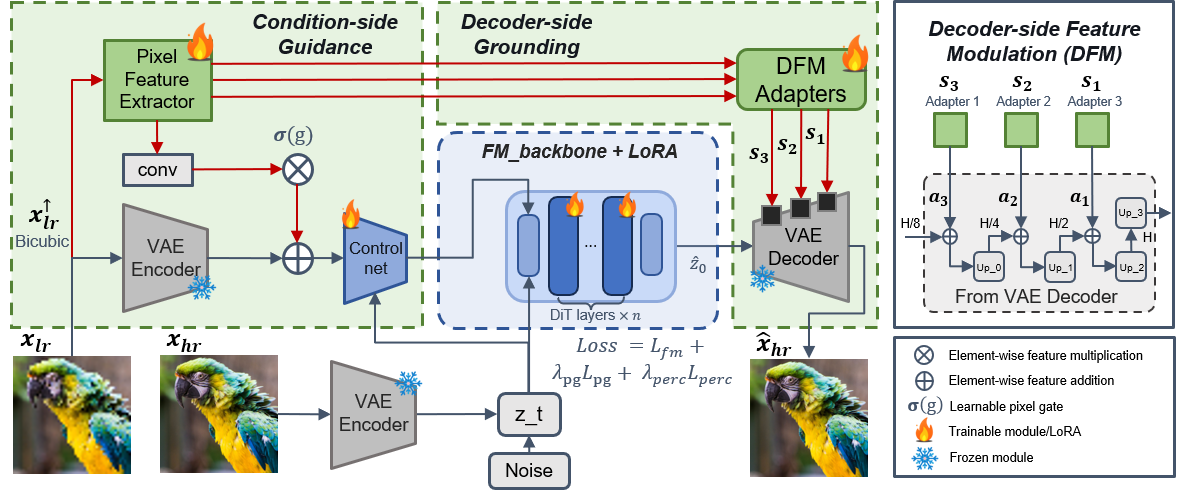}\\
    \caption{
    Overview of the training pipeline of PGSR.
    The LR observation is first aligned to the HR grid and encoded into a latent condition.
    PGSR introduces pixel-grounded evidence at two complementary locations: \yuyao{Condition-Side Trajectory Guidance fuses LR-derived pixel evidence with the latent condition to anchor the flow-matching trajectory, while Decoder-Side Pixel Grounding injects multi-scale pixel features into the frozen VAE decoder for faithful rendering.}
    }
    \label{fig:pipeline}
\end{figure*}

As shown in Fig.~\ref{fig:pipeline}, PGSR reuses pre-VAE pixel evidence at two stages via Pixel-Grounded Guidance: condition-side trajectory guidance anchors the latent flow to LR structure, while decoder-side grounding injects multi-scale cues into the frozen VAE decoder for faithful reconstruction.

We first formulate latent flow-matching SR, then introduce shared pixel evidence with condition-side guidance and decoder-side grounding, followed by the learning objective and a parameter-efficient design.

\subsection{Preliminaries on Latent Flow-Matching Super-Resolution}
\label{sec:fm_prelim}

Let $x_{hr} \in \mathbb{R}^{H \times W \times 3}$ be a high-resolution image and $x_{lr} \in \mathbb{R}^{\frac{H}{s} \times \frac{W}{s} \times 3}$ its low-resolution observation under degradation $\mathcal{D}$ with scale $s$, i.e., $x_{lr}=\mathcal{D}(x_{hr})$.

We upsample $x_{lr}$ to the HR grid, $x_{lr}^{\uparrow}=\mathrm{Bicubic}(x_{lr})$, and encode both images with a frozen VAE:
\[
z_{lr}=E_{\mathrm{vae}}(x_{lr}^{\uparrow}), 
\quad 
z_{hr}=E_{\mathrm{vae}}(x_{hr}).
\]

We learn a conditional velocity predictor in latent space along the flow-matching trajectory. Given $\epsilon\sim\mathcal{N}(0,I)$ and noise level $\sigma_t$,
\[
z_t=\sigma_t\epsilon+(1-\sigma_t)z_{hr}, 
\quad 
v^\star=\epsilon-z_{hr}.
\]
The model predicts $\hat{v}$ conditioned on the LR input and estimates
\[
\hat{z}_0=z_t-\sigma_t\hat{v}, 
\quad 
\mathcal{L}_{\mathrm{fm}}=\|\hat{v}-v^\star\|_2^2.
\]

While efficient, this formulation conditions on the compressed latent $z_{lr}$. We therefore introduce pixel grounding to preserve pre-VAE evidence and guide both the latent trajectory and final decoding.

\subsection{Pixel-Grounded Super-Resolution}

Instead of being a global semantic context, PGSR keeps the LR observation available outside the VAE bottleneck by extracting pixel evidence \textit{before latent encoding}.

\noindent\textbf{Pre-VAE pixel evidence.} Given the upsampled LR image $x_{lr}^{\uparrow}$, a lightweight pixel encoder $P(\cdot)$ produces a latent-resolution feature and multi-scale decoder features:
\[
\bigl(\bar{p}_{lr}, \mathcal{S}_{lr}\bigr)=P(x_{lr}^{\uparrow}), 
\qquad 
\mathcal{S}_{lr}=\{s_1,s_2,s_3\}.
\]
Here $\bar{p}_{lr}$ is used to ground the latent trajectory, while $\{s_1,s_2,s_3\}$ are used to ground the final VAE decoding process.

This pixel evidence does not provide unobserved HR details; instead, it preserves LR-observed cues such as edges, boundaries, and local color transitions before they are compressed by the VAE.

\noindent\textbf{Condition-Side Trajectory Guidance.} The latent LR condition $z_{lr}$ provides a compact structural anchor, while $\bar{p}_{lr}$ supplies complementary pre-VAE image evidence.

\sy{We fuse them through a gated residual correction:
$\tilde{z}_{lr}=z_{lr}+\lambda\,\sigma(g)\,W_p(\bar{p}_{lr})$,
where $W_p$ is a $1\times1$ projection, $g$ is a learnable scalar gate, $\sigma(\cdot)$ is the sigmoid function, and $\lambda$ controls the pixel-conditioning strength.}
The grounded condition is then packed and passed to the ControlNet branch:
\[
\Delta_t=C_\phi(\mathrm{Pack}(z_t),\mathrm{Pack}(\tilde{z}_{lr}),t),
\qquad
\hat{v}=F_{\psi,\omega}(\mathrm{Pack}(z_t),t;\Delta_t),
\]
where $F_{\psi,\omega}$ denotes the pretrained flow-matching transformer with frozen weights $\psi$ and trainable LoRA parameters $\omega$.
Here $\mathrm{Pack}(\cdot)$ denotes the standard rearrangement from latent feature maps to the image-token sequence consumed by the transformer.
This grounds the latent trajectory without replacing the compact latent condition.

\noindent\textbf{Decoder-Side Pixel Grounding.} Condition-Side Trajectory Guidance constrains the predicted latent, but the final image is still rendered by the frozen VAE decoder. To ground this decoding step, PGSR injects the multi-scale pixel features into matched decoder stages through decoder-side feature modulation (DFM). Let $f_d^\ell$ denote the decoder activation before upsampling block $\ell$.
We apply
\[
\tilde{f}_d^\ell=
f_d^\ell + Z_\ell(\mathrm{SiLU}(A_\ell(s_\ell))),
\qquad \ell\in\{0,1,2\},
\]
where $s_\ell$ denotes the scale-matched pixel tap ($s_3$ for $\ell=0$ at $H/8$, $s_2$ for $\ell=1$ at $H/4$, and $s_1$ for $\ell=2$ at $H/2$), $A_\ell$ aligns the pixel feature to the decoder width, and $Z_\ell$ is a zero-initialized output projection.
The final prediction is obtained by $\hat{x}_{hr}
=
D_{\mathrm{vae}}^{\mathrm{pg}}(\hat{z}_0;\mathcal{S}_{lr}).$
Here $D_{\mathrm{vae}}^{\mathrm{pg}}$ denotes the frozen VAE decoder equipped with DFM pixel-grounding adapters.

Because $Z_\ell$ is zero-initialized, the decoder is identical to the frozen VAE at initialization and DFM gradually learns pixel-grounded rendering during training.

\subsection{Efficient Architecture Adaptation}
PGSR is a parameter-efficient adaptation of a pretrained latent flow-matching DiT. We freeze the latent autoencoder and generative backbone, and train only the ControlNet restoration branch, LoRA adapters, pre-VAE pixel encoder, gated trajectory-guidance fusion, and decoder-side pixel-grounding adapters. As SR is purely image-conditioned, we use cached empty-text embeddings and drive restoration solely from the LR input.

For high-resolution efficiency, we adopt local-window attention by replacing dense image-token attention in selected blocks with pretrained local processors:
\[
\mathrm{Attn}(q_i)=\sum_{j\in \mathcal{N}(i)\cup \mathcal{G}}\alpha_{ij}v_j,
\]
where $\mathcal{N}(i)$ is a local neighborhood and $\mathcal{G}$ denotes a set of optional global tokens. PGSR is fine-tuned on this accelerated backbone with the same grounding pathway and objective.

The training objective combines latent flow supervision with pixel-grounded fidelity:
\[
\mathcal{L}
=
\mathcal{L}_{\mathrm{fm}}+\lambda_{\mathrm{perc}}\mathcal{L}_{\mathrm{perc}}+\lambda_{\mathrm{pg}}\mathcal{L}_{\mathrm{pg}},
\]
where $\mathcal{L}_{\mathrm{fm}}$ is the flow-matching loss. Image-space terms are computed from $\hat{x}_{hr}$:
\[
\mathcal{L}_{\mathrm{perc}}=m_{\mathrm{perc}}\,
\mathcal{L}_{\mathrm{LPIPS}}(\hat{x}_{hr},x_{hr}), \quad
\mathcal{L}_{\mathrm{pg}}=\frac{1}{\sum_i m_i}\sum_i m_i
\|\hat{x}_{hr}^{(i)}-x_{hr}^{(i)}\|_1.
\]
Here $m_i=\mathbf{1}[\sigma_i\le\tau]$ is a low-noise mask, with $m_{\mathrm{perc}}$ as its perceptual gate. Image-space supervision is applied only at low noise, where $\hat{z}_0=z_t-\sigma_t\hat{v}$ is stable. Thus, $\mathcal{L}_{\mathrm{fm}}$ guides the full trajectory, while $\mathcal{L}_{\mathrm{perc}}$ and $\mathcal{L}_{\mathrm{pg}}$ refine reliable pixel-grounded outputs.

\section{Experiments}
\paragraph{Datasets.}
We use two groups of training datasets and three evaluation benchmarks.
For clean paired pretraining, we use DF2K, which combines DIV2K \citep{agustsson2017div2k} and Flickr2K \citep{wang2018esrgan}, with paired bicubic $\times4$ LR--HR images.
For real-degradation adaptation, we construct a larger mixed HR corpus from the union of DF2K, LSDIR \citep{li2023lsdir}, FFHQ \citep{karras2019stylegan}, and OST \citep{wang2018esrgan}; LR inputs are synthesized online with a real-world degradation pipeline.
For evaluation, we report results on DIV2K validation \citep{agustsson2017div2k}, RealSR \citep{cai2019realsr}, and DRealSR \citep{wei2020drealsr}, covering both synthetic bicubic degradation and real captured LR--HR pairs.

\paragraph{Training and inference.}
We instantiate PGSR with FLUX.1-dev \citep{blackforestlabs2024fluxdev} and initialize the restoration control branch from a pretrained SR ControlNet compatible with the base flow-matching model \citep{zhang2023controlnet}.
Stage 1 learns the clean SR mapping on DF2K and selects the best checkpoint by DIV2K validation PSNR.
Stage 2 fine-tunes from this checkpoint on the mixed HR corpus using the second-order Real-ESRGAN degradation process \citep{wang2021realesrgan}, switches validation to real-world data, and selects the main model by RealSR validation LPIPS.
In this stage, we enable the decoder-side DFM branch and the low-noise image-space losses while continuing to train the condition-side trajectory guidance modules.
At inference time, PGSR follows the pretrained flow-matching sampling trajectory conditioned on the LR observation.
Given the LR latent $z_{lr}$ and Gaussian noise $\epsilon$, the initial latent is formed as
$z_{t_0}^{\mathrm{init}} = \sigma_{t_0}\epsilon + (1 - \sigma_{t_0}) z_{lr}$,
where $t_0$ controls how strongly the sampler departs from the LR latent condition.
Unless otherwise specified, we use the full restoration trajectory and decode the final latent with the DFM-enabled VAE decoder.

\paragraph{Evaluation metrics.}
We evaluate distortion fidelity with PSNR and SSIM \citep{wang2004ssim}, reference-based perceptual quality with LPIPS \citep{zhang2018lpips} and DISTS \citep{ding2020dists}, and distribution-level fidelity with FID \citep{heusel2017fid}.
We further report no-reference perceptual metrics, including NIQE \citep{mittal2013niqe}, MUSIQ \citep{ke2021musiq}, MANIQA \citep{yang2022maniqa}, and CLIP-IQA \citep{wang2023clipiqa}.
Higher is better for PSNR, SSIM, MUSIQ, MANIQA, and CLIP-IQA; lower is better for LPIPS, DISTS, FID, and NIQE.

\paragraph{Implementation details.}
All experiments are conducted on 8 NVIDIA RTX 6000 Ada GPUs, each with 48GB of memory.
Training uses bf16 mixed precision and distributed data parallelism via Hugging Face Accelerate.
The VAE and pretrained FLUX backbone are frozen; the trainable components are the ControlNet branch, the shared pixel feature extractor, the condition-side trajectory guidance fusion layer, the decoder-side DFM adapters, and FLUX LoRA adapters.
We optimize with AdamW using cosine learning-rate decay and linear warmup. Additional training, adaptation, degradation, and inference details are provided in Appendix~\ref{app:implementation_details}.

\begin{figure*}[t]
\centering
\begin{minipage}[t]{0.6\textwidth}
    \vspace{0pt}
    \centering
    \includegraphics[width=\linewidth]{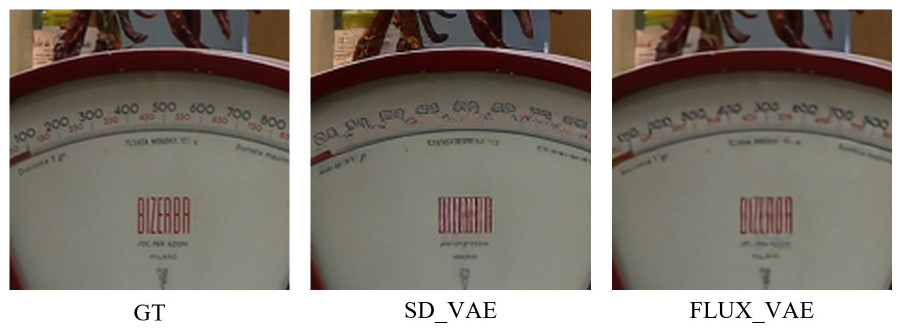}
\end{minipage}
\hfill
\begin{minipage}[t]{0.39\textwidth}
    \vspace{2em}
    \centering
    \setlength{\tabcolsep}{3.2pt}
    \scriptsize
    \begin{tabular}{lccc}
    \toprule
    Method & RMSE$\downarrow$ & PSNR$\uparrow$ & SSIM$\uparrow$ \\
    \midrule
    GT & 0.00 & $\infty$ & 1.0000 \\
    SD VAE & 14.42 & 24.95 & 0.8454 \\
    FLUX VAE & 13.77 & 25.35 & 0.8528 \\
    \bottomrule
    \end{tabular}
\end{minipage}
\vspace{-0.1in}
\caption{VAE reconstruction stress test on an example. Left: visual comparison under controlled latent reconstruction. Right: quantitative reconstruction metrics. Both Stable Diffusion VAE and FLUX VAE exhibit substantial information loss due to latent compression, leading not only to blurring but also to severe structural degradation, where characters, digits, and fine patterns become largely unreadable or distorted beyond recognition.}
\vspace{-0.15in}
\label{fig:vae_stress_test}
\end{figure*}

\paragraph{Necessity of pixel-level information for decoder-side grounding.}
We conduct a controlled VAE reconstruction experiment to analyze the fidelity of latent compression in standard encoder-decoder pipelines. Specifically, we feed ground-truth HR images into the VAE and reconstruct them via the standard encode-decode process, using both the Stable Diffusion VAE \citep{rombach2022ldm} and FLUX VAE \citep{blackforestlabs2024fluxdev}. As shown in Fig.~\ref{fig:vae_stress_test}, both models exhibit noticeable reconstruction degradation, including distorted patterns and blurred fine-grained details, indicating that the VAE bottleneck is inherently lossy even under perfect input conditions. This suggests that encoder-decoder-based SR pipelines do not fully preserve or utilize pixel-level signals during reconstruction, resulting in unavoidable degradation of high-frequency content. 
These results highlight an important limitation of latent compression in existing generative SR systems: important spatial details are partially discarded during encoding, and cannot be fully recovered by the decoder alone.

\subsection{Comparison with State-of-the-Art Methods}

\paragraph{Compared methods.}
We compare PGSR with representative generative SR methods which are reproducible, including ResShift \citep{yue2023resshift}, StableSR \citep{wang2023stablesr}, DiffBIR \citep{lin2024diffbir}, SeeSR \citep{wu2023seesr}, PASD \citep{yang2023pasd}, OSEDiff \citep{wu2024osediff}, SinSR \citep{wang2023sinsr}, and PiSA-SR \citep{sun2025pisasr}.

\begin{figure*}[t]
    \centering
    \includegraphics[width=\textwidth]{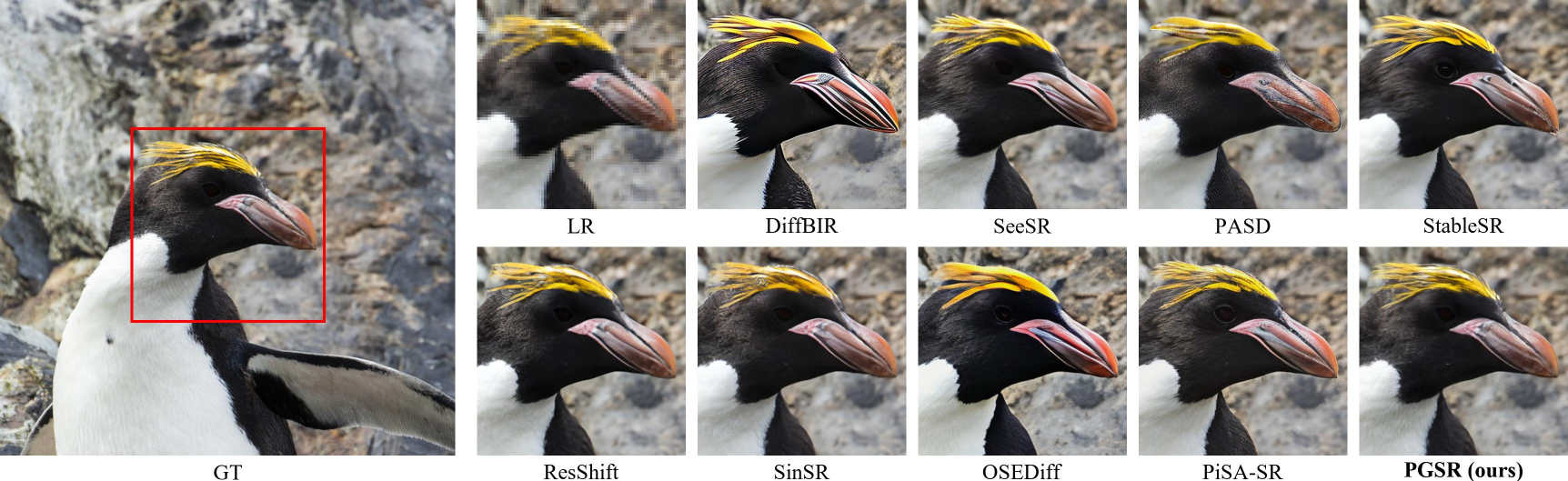}\\ [0.3em]
    \includegraphics[width=\textwidth]{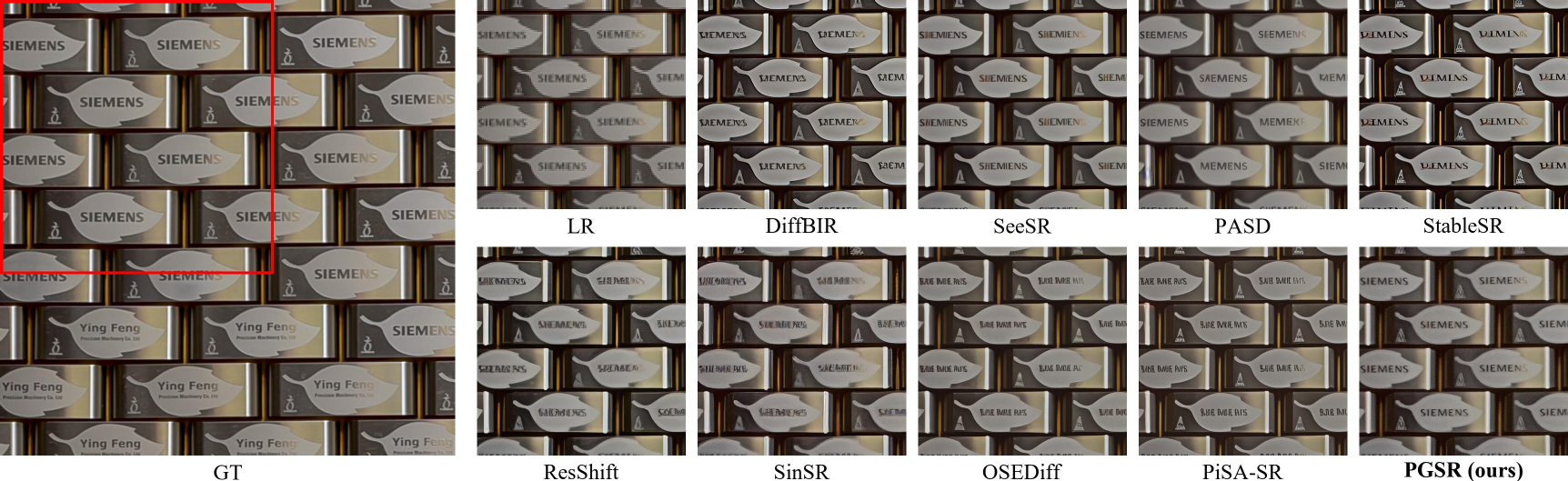}
    \caption{
    Qualitative comparison on representative $\times4$ SR examples from both synthetic bicubic degradation and real degradation.
    PGSR aims to preserve LR-consistent structures while recovering perceptually plausible high-frequency details, including fine feather structures and text characters.
    }
    \label{fig:visual_comparison}
\end{figure*}

\paragraph{Qualitative comparisons.}
Figure~\ref{fig:visual_comparison} shows representative visual comparisons on both synthetic bicubic and real-degradation benchmarks.
Compared with prior generative SR methods, PGSR is designed to preserve LR-consistent structures while recovering plausible high-frequency details, especially around repeated textures and small semantic regions.
The first row presents a penguin image from the DIV2K-Val dataset with virtual bicubic degradation, where the competing methods either produce overly sharp hair and hallucinated eye details or change the saturation of the image.
In contrast, PGSR better preserves the spatial layout inherited from the LR input while producing sharp local details, leading to a more consistent result with the ground-truth.
The second row shows a RealSR captured image, where the degradation is real-world degradation rather than a simple bicubic downsampling process. Baselines such as StableSR and DiffBIR hallucinate textures or distort small structures, whereas PGSR preserves object boundaries and correctly renders the text.
\begin{table*}[h]
\centering
\caption{Quantitative comparison with representative generative SR methods on synthetic and real-world $\times4$ benchmarks.
All methods are evaluated using the same local metric pipeline, covering distortion metrics, full-reference perceptual metrics, distribution-level fidelity, and no-reference perceptual quality.
For each dataset and metric, the best, second-best, and third-best results are highlighted in \textbf{bold}, \underline{underline}, and \textit{italics}, respectively.}
\label{tab:main_results}
\setlength{\tabcolsep}{3.2pt}
\scriptsize
\resizebox{\textwidth}{!}{%
\begin{tabular}{l|l|ccccc|cccc}
\toprule
\textbf{Dataset} & \textbf{Method} & PSNR$\uparrow$ & SSIM$\uparrow$ & LPIPS$\downarrow$ & DISTS$\downarrow$ & FID$\downarrow$
& NIQE$\downarrow$ & MUSIQ$\uparrow$ & MANIQA$\uparrow$ & CLIP-IQA$\uparrow$ \\
\midrule
\multirow{9}{*}{DIV2K}
& ResShift & \textit{25.56} & \textit{0.7625} & 0.2166 & \underline{0.0797} & \textit{28.36} & 4.8223 & 62.74 & 0.3555 & 0.6276 \\
& StableSR & 23.55 & 0.7050 & 0.2272 & 0.0850 & 32.50 & 3.7174 & \textit{68.55} & \underline{0.4591} & \textit{0.7080} \\
& DiffBIR & 23.04 & 0.6852 & 0.2731 & 0.1054 & 36.48 & 3.6071 & 68.06 & \textbf{0.4639} & \textbf{0.7370} \\
& SeeSR & 25.51 & 0.7566 & 0.2225 & 0.0866 & 31.19 & 3.8600 & 67.11 & 0.4340 & 0.6403 \\
& PASD & 24.86 & 0.7373 & 0.2361 & \textit{0.0849} & 29.94 & \textit{3.3494} & 66.66 & 0.3975 & 0.5996 \\
& OSEDiff & 23.16 & 0.7058 & 0.2386 & 0.1031 & 34.30 & 3.4451 & \underline{69.34} & 0.4404 & 0.7004 \\
& SinSR & \underline{25.78} & \textbf{0.7781} & \underline{0.2107} & 0.0862 & \underline{27.39} & 4.2933 & 64.58 & 0.3881 & 0.6582 \\
& PiSA-SR & 23.59 & 0.7087 & \textit{0.2158} & 0.0937 & 32.71 & \underline{3.2426} & \textbf{70.46} & \textit{0.4556} & \underline{0.7190} \\
\cmidrule{2-11}
& \textbf{PGSR (ours)} & \textbf{25.89} & \underline{0.7707} & \textbf{0.2104} & \textbf{0.0792} & \textbf{24.88} & \textbf{3.2244} & 67.93 & 0.3913 & 0.6479 \\
\midrule
\multirow{9}{*}{DRealSR}
& ResShift & 24.73 & 0.6702 & 0.4298 & 0.1078 & 54.11 & 6.6914 & 31.91 & 0.3060 & 0.6069 \\
& StableSR & 24.77 & 0.6934 & 0.4194 & 0.1116 & 55.07 & 4.5860 & \textit{34.78} & 0.4094 & 0.6720 \\
& DiffBIR & 24.83 & 0.5979 & 0.4582 & 0.1255 & 54.30 & 4.6915 & 34.75 & \textit{0.4132} & 0.6890 \\
& SeeSR & \underline{25.91} & \textbf{0.7755} & \textbf{0.2761} & \underline{0.0962} & \textit{46.91} & 5.3388 & 33.33 & 0.3606 & 0.5459 \\
& PASD & \textit{25.84} & 0.7464 & \underline{0.2898} & 0.1087 & 50.28 & 6.2480 & 29.54 & 0.3194 & 0.4442 \\
& OSEDiff & 24.69 & 0.7358 & 0.3244 & 0.1117 & \underline{46.61} & \underline{4.1527} & \textbf{37.01} & \textbf{0.4764} & \underline{0.7016} \\
& SinSR & 25.47 & 0.6632 & 0.4496 & \textit{0.1047} & 60.62 & 5.7972 & 31.33 & 0.3590 & \textit{0.6937} \\
& PiSA-SR & 25.37 & \underline{0.7592} & 0.3133 & 0.1142 & 49.88 & \textit{4.4010} & 34.67 & \underline{0.4456} & \textbf{0.7171} \\
\cmidrule{2-11}
& \textbf{PGSR (ours)} & \textbf{25.97} & \textit{0.7550} & \textit{0.3054} & \textbf{0.0911} & \textbf{46.15} & \textbf{3.8707} & \underline{35.83} & 0.3144 & 0.5932 \\
\midrule
\multirow{9}{*}{RealSR}
& ResShift & 24.06 & 0.7112 & 0.3491 & 0.1758 & 63.61 & 6.9271 & 55.20 & 0.3336 & 0.5647 \\
& StableSR & 22.77 & 0.6974 & 0.3288 & 0.1748 & 63.08 & 5.0611 & 62.57 & 0.4446 & 0.6284 \\
& DiffBIR & 23.74 & 0.6353 & 0.3456 & 0.1807 & 59.47 & 4.9181 & 62.51 & \textit{0.4700} & \textbf{0.6958} \\
& SeeSR & \underline{25.13} & \textbf{0.7620} & 0.2750 & \textit{0.1553} & 61.99 & 5.3505 & 61.20 & 0.4380 & 0.5984 \\
& PASD & \textbf{25.20} & \textit{0.7477} & \textit{0.2737} & \underline{0.1473} & \textbf{52.85} & 4.8421 & 58.48 & 0.3741 & 0.5182 \\
& OSEDiff & 23.68 & 0.7163 & 0.2999 & 0.1609 & \textit{58.80} & \underline{4.3153} & \underline{67.29} & \textbf{0.4758} & \underline{0.6832} \\
& SinSR & 24.72 & 0.7080 & 0.3642 & 0.1766 & 67.05 & 5.8306 & 57.86 & 0.3844 & 0.6405 \\
& PiSA-SR & 24.03 & 0.7307 & \underline{0.2719} & 0.1564 & 61.42 & \textit{4.3685} & \textbf{67.97} & \underline{0.4746} & \textit{0.6745} \\
\cmidrule{2-11}
& \textbf{PGSR (ours)} & \textit{24.75} & \underline{0.7495} & \textbf{0.2596} & \textbf{0.1365} & \underline{54.63} & \textbf{4.0470} & \textit{63.39} & 0.4116 & 0.5593 \\
\bottomrule
\end{tabular}%
}
\vspace{-0.1in}
\end{table*}

\begin{table*}[h]
\centering
\caption{Architecture ablation of pixel-grounded pathways in PGSR.
The left table reports quantitative results on DIV2K-Val and RealSR, while the right panel shows a representative visual ablation.
Removing either the condition-side trajectory guidance or the decoder-side grounding degrades reconstruction and perceptual quality, and the full design gives the best results on both datasets.}
\label{tab:arch_ablation}
\begin{minipage}[t]{0.4\textwidth}
\vspace{1em}
\setlength{\tabcolsep}{3.0pt}
\scriptsize
\resizebox{\linewidth}{!}{%
\begin{tabular}{l l c c c}
\toprule
Dataset & Variant & PSNR$\uparrow$ & SSIM$\uparrow$ & LPIPS$\downarrow$ \\
\midrule
\multirow{4}{*}{DIV2K}
& w/o Pixel-Grounded & 23.68 & 0.7592 & 0.2431 \\
& w/o Condition-Side & 24.66 & 0.7644 & 0.2204 \\
& w/o Decoder-Side & 24.78 & 0.7651 & 0.2188 \\
& \textbf{PGSR (full)} & \textbf{25.89} & \textbf{0.7707} & \textbf{0.2104} \\
\midrule
\multirow{4}{*}{RealSR}
& w/o Pixel-Grounded & 22.15 & 0.7305 & 0.2789 \\
& w/o Condition-Side & 24.22 & 0.7420 & 0.2631 \\
& w/o Decoder-Side & 24.08 & 0.7401 & 0.2626 \\
& \textbf{PGSR (full)} & \textbf{24.75} & \textbf{0.7495} & \textbf{0.2596} \\
\bottomrule
\end{tabular}
}
\end{minipage}
\hfill
\begin{minipage}[t]{0.59\textwidth}
\vspace{0pt}
\centering
\includegraphics[width=\linewidth]{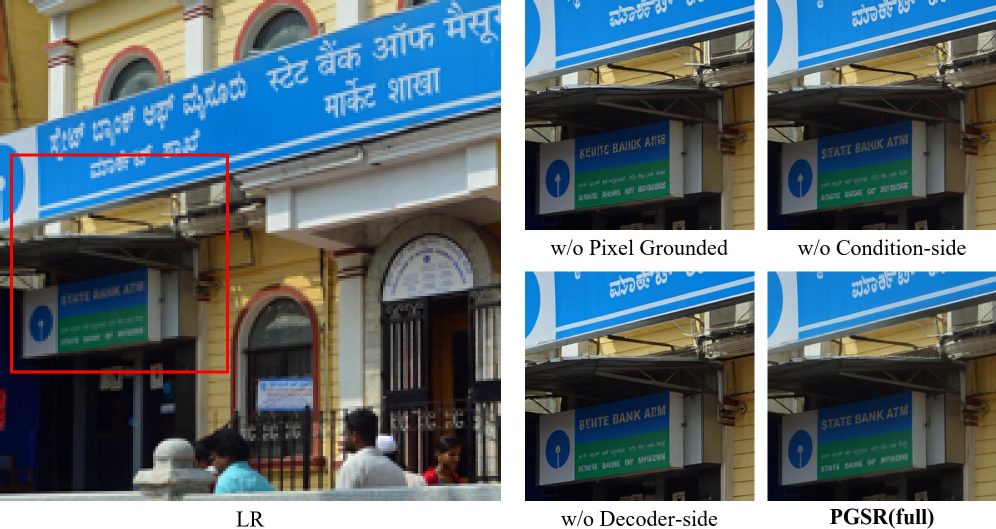}
\vspace{-0.1in}
\end{minipage}
\vspace{-0.15in}
\end{table*}

\paragraph{Quantitative comparisons.}
\yuyao{Table~\ref{tab:main_results} reports quantitative comparisons on synthetic and real-world $\times4$ SR benchmarks using distortion, perceptual, distribution-level, and no-reference quality metrics.
On DIV2K-Val, PGSR achieves the best PSNR, LPIPS, DISTS, FID, and NIQE, and ranks second in SSIM, indicating that preserving pre-VAE pixel evidence improves fidelity to the LR observation while maintaining strong perceptual quality.}
\yuyao{Notably, compared with perception-oriented methods such as DiffBIR and PiSA-SR, PGSR obtains substantially better distortion and perceptual metrics, suggesting that Pixel-Grounded Guidance reduces visually plausible but input-inconsistent hallucinations.}
\yuyao{On DRealSR, PGSR obtains the best PSNR, DISTS, FID, and NIQE, showing strong robustness to real-world degradations and degradation traces.
On RealSR, PGSR achieves the best LPIPS, DISTS, and NIQE, and ranks second in SSIM and FID.
These results demonstrate a favorable realism--fidelity trade-off across both synthetic and real settings: the latent flow-matching backbone provides realistic textures, while trajectory guidance and decoder-side grounding constrain the restoration with LR-observed pixel evidence during latent generation and final rendering.}

\subsection{Ablation Study}
\paragraph{Ablation study on PGSR components.}
Table~\ref{tab:arch_ablation} shows that removing either grounding pathway degrades PSNR, SSIM, and LPIPS, while removing both causes the largest drop. The full model performs best on both datasets, supporting complementary trajectory and decoder grounding. Loss ablations are provided in Appendix~\ref{app:loss_ablation}.

\begin{figure}[h]
    \centering
    \includegraphics[width=\linewidth]{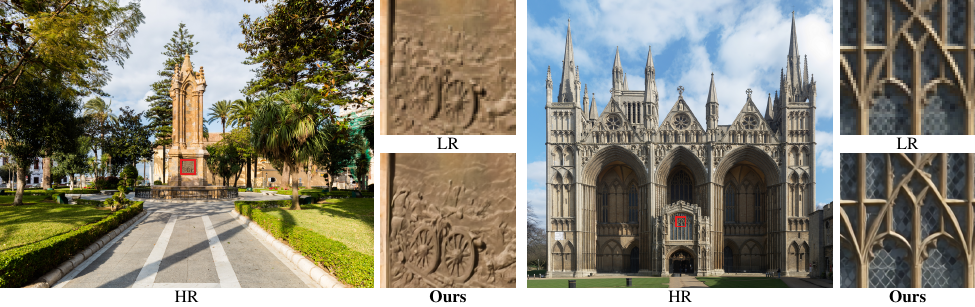}
    \vspace{-0.1in}
    \caption{8K Super-resolution demo using our method. The super-resolution results demonstrate clear quality improvements over the LR input on the DIV8K dataset.}
    \label{fig:8K}
    \vspace{-0.1in}
\end{figure}

\paragraph{Impact of Sparse Attention.}
At fixed output sizes of $1792^2$ (DIV2K-Val) and $1536^2$ (RealSR), sparse attention reduces inference time by approximately 37\% with a modest quality decrease (Table~\ref{tab:clear_efficiency}, left). This benchmark is separate from Table~\ref{tab:main_results}. Sparse attention also enables $2\mathrm{K}\rightarrow8\mathrm{K}$ SR (Fig.~\ref{fig:8K}); Appendix~\ref{app:grounding_flops} reports pixel-grounding FLOPs.

\paragraph{Text preservation.}
Using PARSeq~\citep{bautista2022parseq}, we evaluate 1,339 words (4--8 LR pixels high) in 100 TextOCR-val images~\citep{singh2021textocr} with bicubic $\times4$ degradation, fixed GT quadrilateral crops, and lowercase alphanumeric normalization. PGSR achieves 27.48\% word accuracy and 57.33\% character error rate (CER) (Table~\ref{tab:ocr_preservation}, right).

\begin{table*}[h]
\centering
\vspace{-1em}
\caption{Left: efficiency--quality comparison of dense and sparse PGSR with 20 steps. Right: small-text OCR; HR calibrates recognizer error. Bold: best restored-input result.}
\label{tab:clear_efficiency}
\label{tab:ocr_preservation}
\begin{minipage}[t]{0.66\textwidth}
\vspace{0pt}
\centering
\setlength{\tabcolsep}{2pt}
\footnotesize
\renewcommand{\arraystretch}{1}
\begin{tabular*}{\linewidth}{@{\extracolsep{\fill}}llrrrrr@{}}
\toprule
\multirow{2}{*}{Dataset} & \multirow{2}{*}{Attention} & \multicolumn{2}{c}{Efficiency} & \multicolumn{3}{c}{Quality} \\
\cmidrule(lr){3-4}\cmidrule(l){5-7}
& & s/img$\downarrow$ & s/step$\downarrow$ & PSNR$\uparrow$ & SSIM$\uparrow$ & LPIPS$\downarrow$ \\
\midrule
\multirow{2}{*}{DIV2K}
& Full & 61.81 & 3.09 & 26.36 & 0.7710 & 0.2571 \\
& Sparse & 38.95 & 1.95 & 25.81 & 0.7679 & 0.2603 \\
\midrule
\multirow{2}{*}{RealSR}
& Full & 58.28 & 2.91 & 24.88 & 0.7501 & 0.2582 \\
& Sparse & 37.09 & 1.85 & 24.37 & 0.7482 & 0.2601 \\
\bottomrule
\end{tabular*}
\end{minipage}\hfill
\begin{minipage}[t]{0.32\textwidth}
\vspace{0pt}
\centering
\footnotesize
\setlength{\tabcolsep}{2pt}
\renewcommand{\arraystretch}{1}
\begin{tabular}{lrr}
\toprule
Method & Acc.\ (\%)$\uparrow$ & CER (\%)$\downarrow$ \\
\noalign{\vskip\dimexpr\aboverulesep+\belowrulesep+\cmidrulewidth\relax}
\midrule
HR reference & 86.41 & 4.97 \\
\midrule
PASD & 8.36 & 82.00 \\
SeeSR & 7.17 & 85.77 \\
PiSA-SR & 7.32 & 82.40 \\
\textbf{PGSR} & \textbf{27.48} & \textbf{57.33} \\
\bottomrule
\end{tabular}
\end{minipage}
\end{table*}

\vspace{-1.2em}

\paragraph{Limitations.} 

PGSR inherits both the strengths and costs of large pretrained flow-matching models. Due to the large ControlNet and the backbone, training requires substantial memory, and inference remains slower than one-step SR models.
Besides, since our losses emphasize LR consistency and a mean-error learning objective, PGSR can produce slightly smoother textures than methods that aggressively optimize perceptual sharpness.
This reflects a realism--fidelity trade-off; future work may improve sharpness through distillation while preserving input-output consistency.

\section{Conclusion}

We presented PGSR, a generative super-resolution framework centered on Pixel-Grounded Guidance.
Rather than treating the LR image only as a latent condition, PGSR reuses pre-VAE pixel evidence through two complementary mechanisms: Condition-Side Trajectory Guidance anchors the ControlNet condition and latent flow trajectory, while Decoder-Side Pixel Grounding injects multi-scale pixel evidence into the frozen VAE decoder.
This dual design directly addresses the fidelity loss caused by latent compression and provides a practical path toward more faithful generative SR with large pretrained diffusion transformers.

\bibliographystyle{plainnat}
\bibliography{reference}

\clearpage
\appendix
\section{Additional Implementation Details}
\label{app:implementation_details}
We provide additional implementation details to facilitate reproducibility. Unless otherwise stated, all reported PGSR models are initialized from \texttt{black-forest-labs/FLUX.1-dev}. The restoration ControlNet branch is initialized from a FLUX-compatible SR ControlNet checkpoint, while the FLUX backbone and the VAE encoder--decoder remain frozen throughout training. The trainable components include the ControlNet restoration branch, the pre-VAE pixel extractor, the gated condition-side pixel fusion layer, decoder-side DFM adapters, and LoRA adapters inserted into the FLUX backbone.

\paragraph{Optimization.}
We train PGSR with AdamW using a base learning rate of $1\times 10^{-5}$, weight decay $0.01$, and Adam coefficients $(\beta_1,\beta_2)=(0.9,0.999)$. The learning rate is linearly warmed up for the first 5 epochs and then decayed with a cosine schedule. The ControlNet branch uses the base learning rate, while the pixel branch and LoRA parameters use a higher learning rate of $1\times 10^{-4}$. Training is performed with bfloat16 mixed precision on 8 NVIDIA RTX 6000 Ada GPUs. Unless otherwise specified, we use a per-GPU batch size of 4 and gradient accumulation of 8, resulting in an effective batch size of 256. We use a fixed random seed of 42 for the reported runs.

\paragraph{Training protocol.}
For the clean paired stage, we train on DF2K with paired bicubic $\times 4$ LR--HR images. HR crops of size $512\times512$ are used, corresponding to $128\times128$ LR crops for $\times 4$ SR. We sample two crops per image and train for 60 epochs. The clean-stage checkpoint is selected according to DIV2K validation PSNR. For the real-world degradation stage, we fine-tune the clean checkpoint using a Real-ESRGAN-style second-order degradation process on the mixed HR corpus described in Sec.~4. The real-world stage uses validation on real captured data, and the checkpoint used for the main real-world results is selected according to RealSR validation LPIPS.

\paragraph{Model adaptation.}
The condition-side pixel feature is fused with the LR latent condition through the gated residual formulation described in Sec.~3.2. The pixel gate is initialized with a positive logit, so that pixel evidence is active at the beginning of training while still being learnable. LoRA adapters use rank 16, alpha 16, zero dropout, and Gaussian initialization. Since SR is fully image-conditioned, we use cached empty-text embeddings during training and inference. In our implementation, we additionally allow a small set of learnable leading text tokens with a substantially smaller learning rate and L2 regularization; this component is used only as a lightweight prior adapter and is not the main source of image conditioning. 

\paragraph{Decoder-side pixel grounding.}
The decoder-side DFM branch is enabled for the full PGSR model. Multi-scale pixel features are injected into the frozen VAE decoder at three matched resolutions: the deepest pixel tap is injected into the first upsampling stage, the intermediate tap into the second stage, and the shallowest tap into the third stage. The pixel extractor uses progressively wider channels across scales, and the latent-resolution pixel feature is projected to match the latent conditioning width before gated fusion. For image-space supervision, we apply the low-noise gating described in Sec.~3.3, so that pixel and perceptual losses are only applied when the predicted clean latent is sufficiently stable.

\paragraph{Degradation settings.}
In the clean paired stage, the model is trained with paired bicubic $\times4$ LR--HR data rather than online real-world degradation. In the real-world stage, we use a second-order degradation pipeline following Real-ESRGAN. The pipeline includes stochastic blur, resizing, Gaussian or Poisson noise, JPEG compression, optional second degradation, final sinc filtering, and unsharp masking of the HR target. We use this stage to expose PGSR to realistic blur--noise--compression artifacts while preserving the same pixel-grounded architecture.

\paragraph{Inference.}
Unless otherwise specified, PGSR uses 20 denoising steps at inference time. We use 20 denoising steps and set the denoising strength to 1.0, i.e., sampling starts from the noisiest restoration state and follows the full restoration trajectory. The final latent prediction is decoded with the DFM-enabled VAE decoder. For large images, we use tiled evaluation with a default tile size of 512 and overlap of 64 to reduce memory usage while avoiding visible boundary artifacts.

\section{Supplementary Experiments}
\label{app:supp_experiments}
\begin{figure*}[p]
    \centering
    \includegraphics[width=\textwidth]{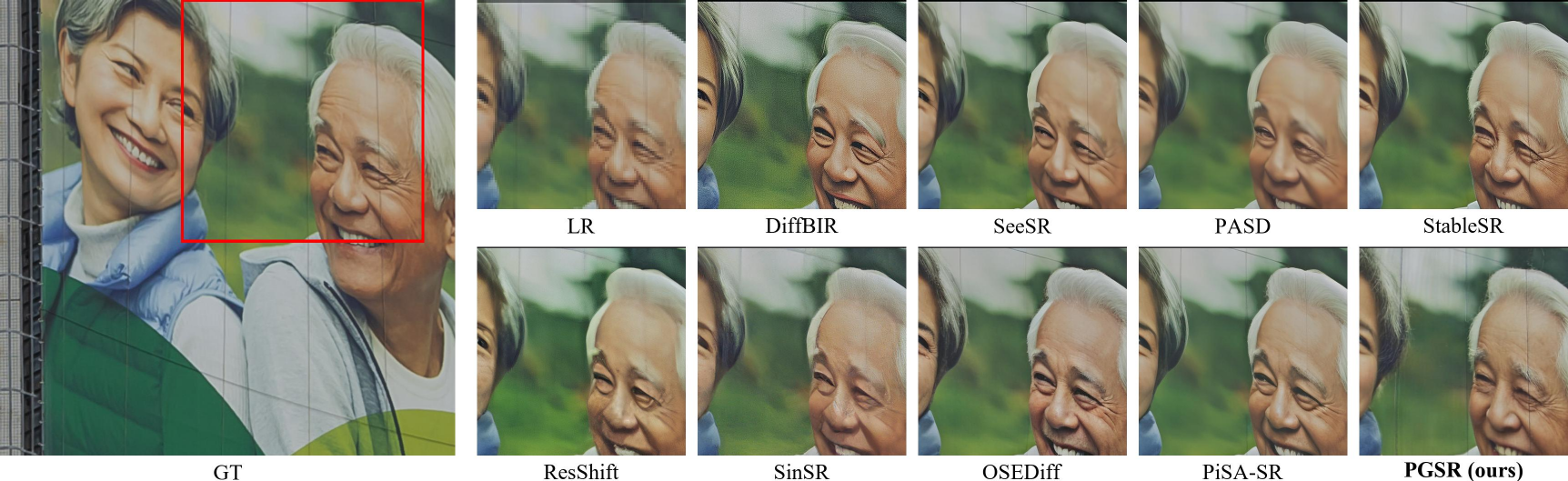}\\ [0.3em]
    \includegraphics[width=\textwidth]{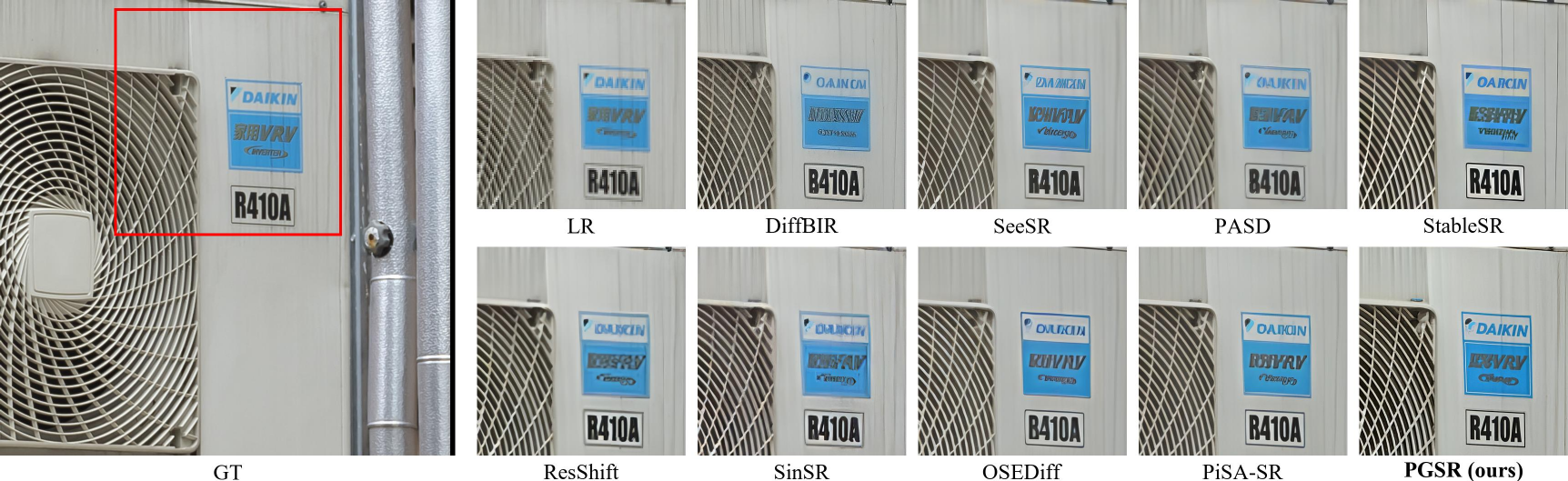}\\ [0.3em]
    \includegraphics[width=\textwidth]{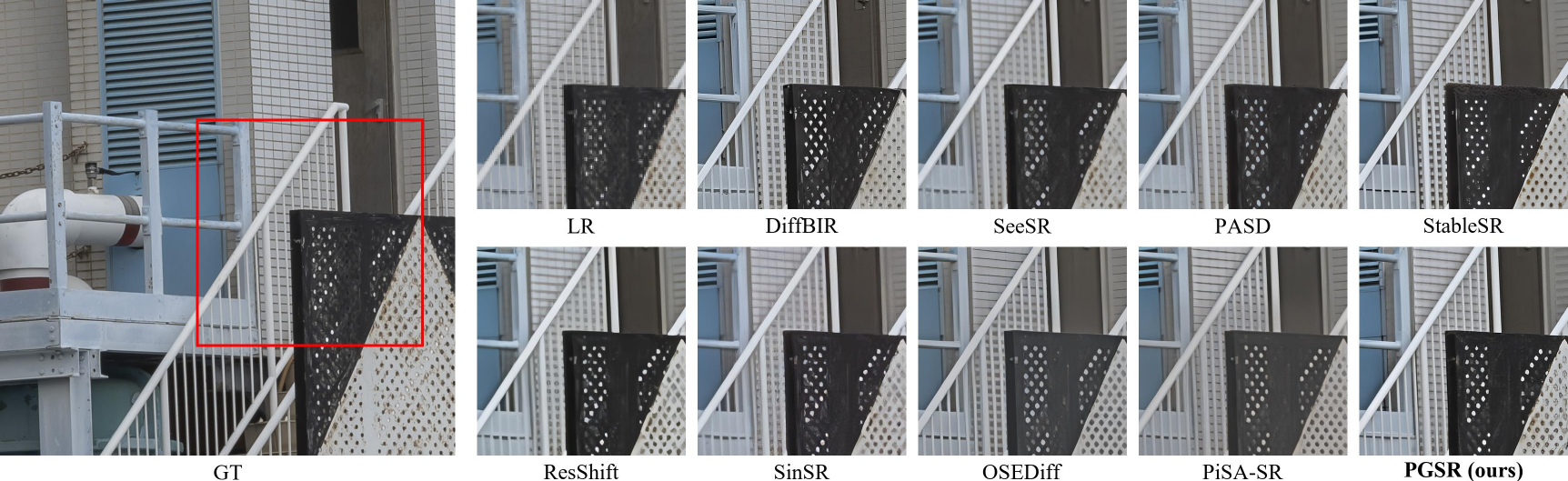}\\ [0.3em]
     \includegraphics[width=\textwidth]{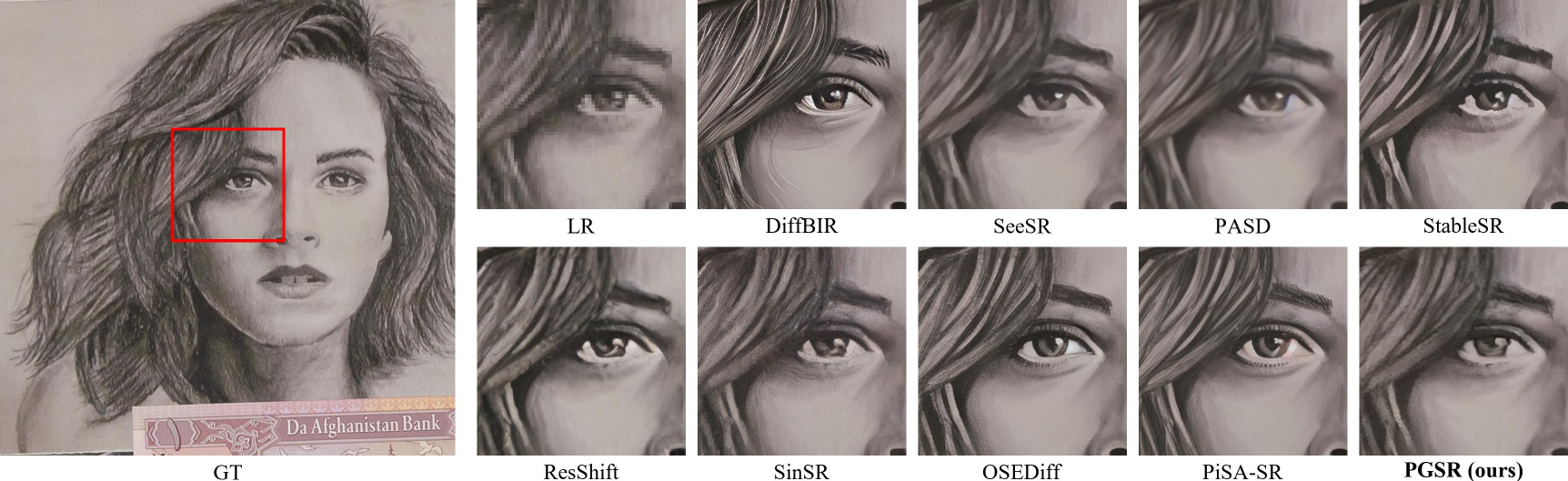}
    \caption{Additional qualitative comparison on real degradation.}
    \label{fig:app_vis_realsr}
\end{figure*}

\begin{figure*}[p]
    \centering
    \includegraphics[width=\textwidth]{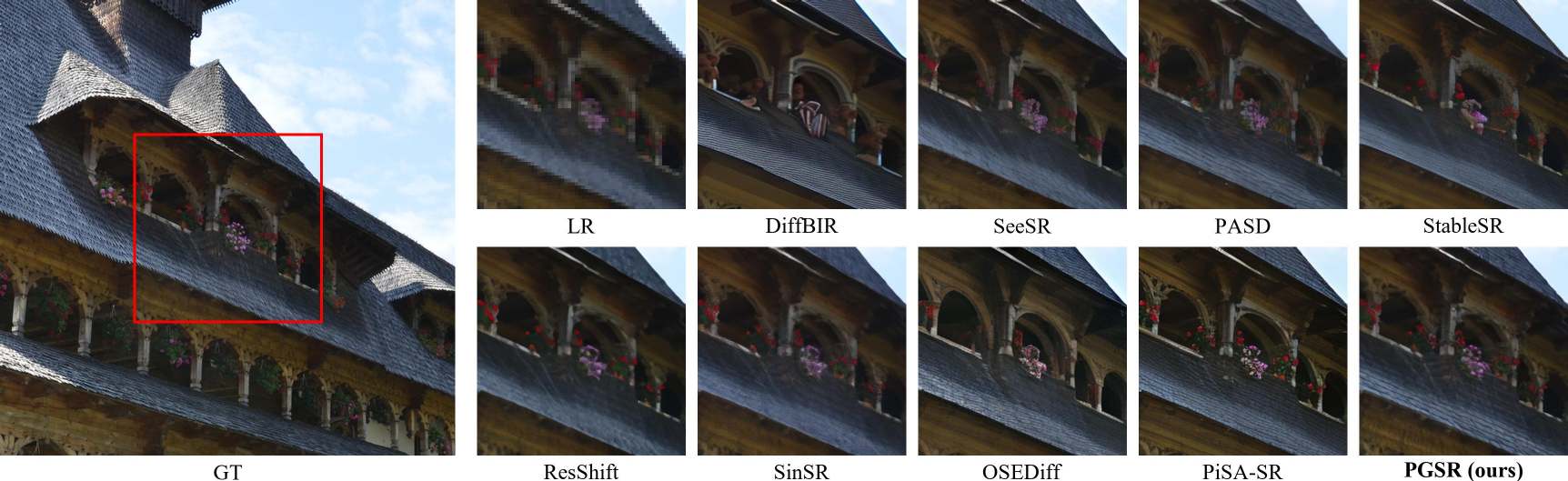}\\ [0.3em]
    \includegraphics[width=\textwidth]{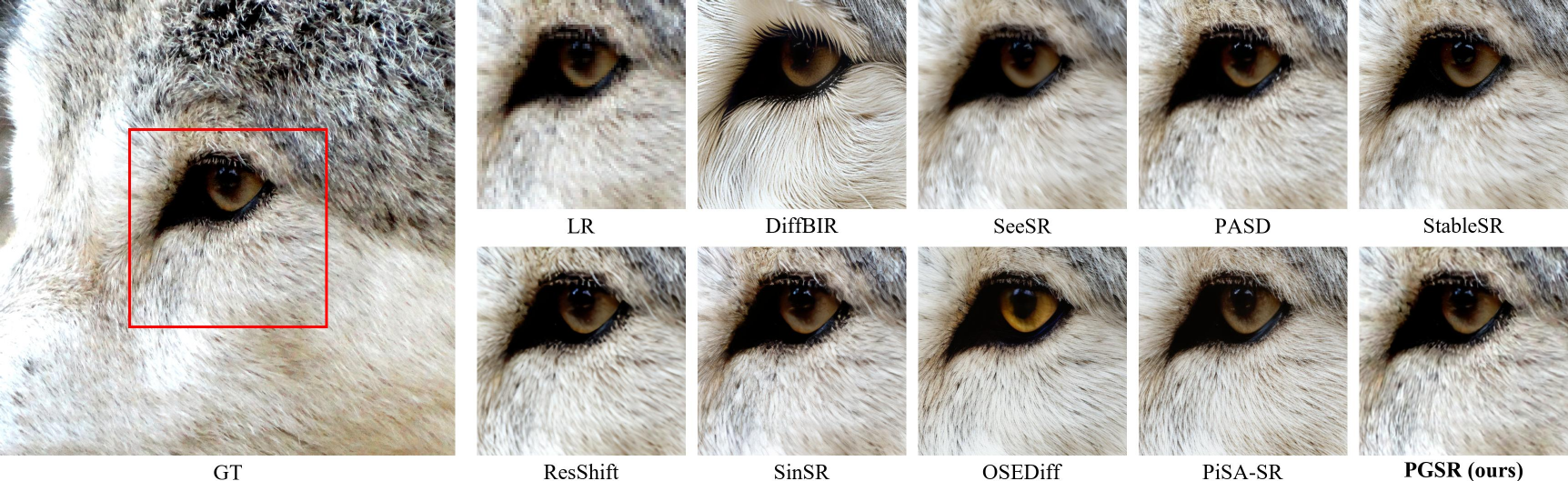}\\ [0.3em]
    \includegraphics[width=\textwidth]{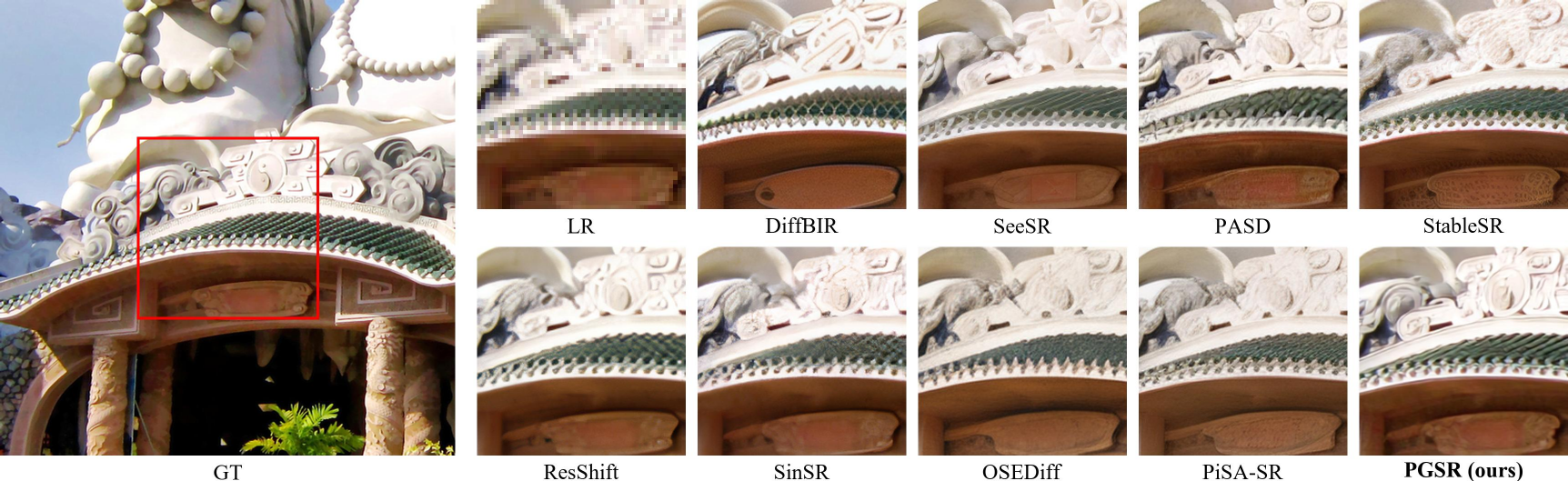}\\[0.3em]
    \includegraphics[width=\textwidth]{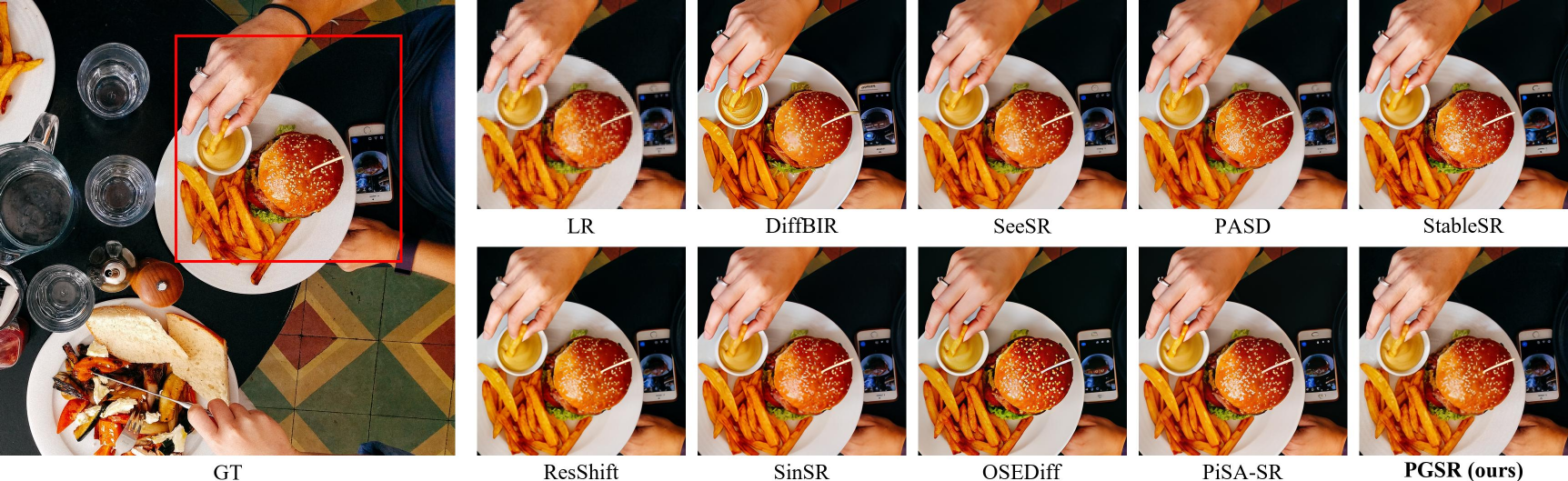}
    \caption{Additional qualitative comparison on synthetic degradation.}
    \label{fig:app_vis_div2k}
\end{figure*}
\subsection{Additional Visual Comparisons}
\label{app:visual_comparisons}

We provide additional qualitative comparisons on RealSR, DRealSR, and DIV2K-Val in Figs.~\ref{fig:app_vis_realsr}--\ref{fig:app_vis_div2k}.
These examples complement the main-text visualization by covering more diverse textures, structures, and real-world degradations.





\subsection{Additional Ablation Tables}
\label{app:loss_ablation}
In this section, we present additional ablation studies on key hyperparameters and loss components.

Table~\ref{tab:hparam_sensitivity_appendix} studies the most relevant training-time hyperparameters around the final Stage-2 recipe.
We vary the perceptual loss weight $\lambda_{\mathrm{perc}}$, the pixel-grounded reconstruction weight $\lambda_{\mathrm{pg}}$, and the pixel-conditioning strength $s_{\mathrm{pc}}$.
Here $s_{\mathrm{pc}}$ scales the condition-side pixel evidence before it is fused into the LR latent condition; it is a conditioning strength and is distinct from the LR-anchored sampling strength used at inference time, which controls the denoising start point.
Other optimizer and engineering settings are kept in the released configuration files.

\begin{table*}[h]
\centering
\caption{Sensitivity to the two image-space loss weights and the pixel-conditioning strength.
Rows are grouped by the factor being swept; within each group, all other settings are fixed to the default Stage-2 recipe.
We report PSNR and LPIPS to show the fidelity--perception effect of each hyperparameter.}
\label{tab:hparam_sensitivity_appendix}
\setlength{\tabcolsep}{3.4pt}
\scriptsize
\resizebox{\textwidth}{!}{%
\begin{tabular}{l l | c c c | c c | c c}
\toprule
\multirow{2}{*}{Sweep} &
\multirow{2}{*}{Variant} &
\multirow{2}{*}{$\lambda_{\mathrm{perc}}$} &
\multirow{2}{*}{$\lambda_{\mathrm{pg}}$} &
\multirow{2}{*}{$s_{\mathrm{pc}}$} &
\multicolumn{2}{c|}{DIV2K-Val} &
\multicolumn{2}{c}{RealSR} \\
\cmidrule(lr){6-7}\cmidrule(lr){8-9}
& & & & & PSNR$\uparrow$ & LPIPS$\downarrow$ & PSNR$\uparrow$ & LPIPS$\downarrow$ \\
\midrule
Loss weights & No image-space loss & 0 & 0 & 1.0 & 24.46 & 0.2301 & 23.25 & 0.2784 \\
Loss weights & Perceptual only & 0.05 & 0 & 1.0 & 24.68 & 0.2172 & 24.01 & 0.2635 \\
Loss weights & Weak pixel loss & 0.05 & 0.025 & 1.0 & 25.06 & 0.2113 & 24.53 & 0.2619\\
Loss weights & \textbf{Default} & \textbf{0.05} & \textbf{0.05} & \textbf{1.0} &\textbf{25.89} & \textbf{0.2104} & \textbf{24.75}& \textbf{0.2596} \\
Loss weights & Strong pixel loss & 0.05 & 0.10 & 1.0 & 25.83 & 0.2160 & 24.32 & 0.2655 \\
\midrule
Conditioning & Weak pixel conditioning & 0.05 & 0.05 & 0.8 & 26.01 & 0.2307 & 24.67 & 0.2706 \\
Conditioning & Strong pixel conditioning & 0.05 & 0.05 & 1.2 & 25.42 & 0.2244 & 24.38 & 0.2664 \\
\bottomrule
\end{tabular}
}
\end{table*}

Table~\ref{tab:loss_ablation_appendix} reports the loss-component ablation used to separate supervision effects from the architectural ablation in the main text.
Starting from the latent flow-matching objective, adding perceptual supervision improves LPIPS on both DIV2K-Val and RealSR, while the pixel-grounded reconstruction loss further improves distortion metrics.
We also test an optional perceptual term on the DFM-decoded prediction to examine whether the decoder-side pixel-grounding branch benefits from perceptual supervision in addition to pixel-wise reconstruction:
\[
\mathcal{L}_{\mathrm{ext}}
=
\mathcal{L}_{\mathrm{fm}}
+\lambda_{\mathrm{perc}}\mathcal{L}_{\mathrm{perc}}
+\lambda_{\mathrm{pg}}\mathcal{L}_{\mathrm{pg}}
+\lambda_{\mathrm{pg\mbox{-}perc}}\mathcal{L}_{\mathrm{pg\mbox{-}perc}},
\quad
\mathcal{L}_{\mathrm{pg\mbox{-}perc}}
=
m_{\mathrm{pg}}\mathcal{L}_{\mathrm{LPIPS}}(\hat{x}_{hr}^{\mathrm{DFM}},x_{hr}).
\]
Here $\hat{x}_{hr}^{\mathrm{DFM}}$ is decoded through the DFM path and $m_{\mathrm{pg}}$ uses the same low-noise gate as $\mathcal{L}_{\mathrm{pg}}$.
The default PGSR recipe sets $\lambda_{\mathrm{pg\mbox{-}perc}}=0$; the final row is included only as an exploratory ablation of perceptual supervision inside the DFM branch.

\begin{table*}[h]
\centering
\caption{Loss-component ablation of PGSR on DIV2K-Val and RealSR validation.
Each row progressively adds one supervision term to the latent flow-matching baseline.
The optional DFM perceptual term tests whether applying LPIPS directly on the DFM-decoded prediction further improves perceptual quality.}
\label{tab:loss_ablation_appendix}
\setlength{\tabcolsep}{3.0pt}
\scriptsize
\resizebox{\textwidth}{!}{%
\begin{tabular}{l c c c c | c c c | c c c}
\toprule
\multirow{2}{*}{Variant} &
\multirow{2}{*}{$\mathcal{L}_{\mathrm{fm}}$} &
\multirow{2}{*}{$\mathcal{L}_{\mathrm{perc}}$} &
\multirow{2}{*}{$\mathcal{L}_{\mathrm{pg}}$} &
\multirow{2}{*}{$\mathcal{L}_{\mathrm{pg\mbox{-}perc}}$} &
\multicolumn{3}{c|}{DIV2K-Val} & \multicolumn{3}{c}{RealSR} \\
\cmidrule(lr){6-8}\cmidrule(lr){9-11}
& & & & & PSNR$\uparrow$ & SSIM$\uparrow$ & LPIPS$\downarrow$ & PSNR$\uparrow$ & SSIM$\uparrow$ & LPIPS$\downarrow$ \\
\midrule
FM control only & \checkmark & -- & -- & -- & 24.67 & 0.7580 & 0.2431 & 22.15 & 0.7305 & 0.2789  \\
+ perceptual & \checkmark & \checkmark & -- & -- & 24.78 & 0.7651 & 0.2188 & 24.08 & 0.7401 & 0.2626 \\
\textbf{PGSR (default)} & \checkmark & \checkmark & \checkmark & -- & \textbf{25.89} & \textbf{0.7707} & \textbf{0.2104} & \textbf{24.75} & 0.7495 & 0.2596 \\
+ DFM perceptual & \checkmark & \checkmark & \checkmark & \checkmark & 25.70 &  0.7696 & \textbf{0.2102} & 24.36 & 0.7490 & \textbf{0.2588} \\
\bottomrule
\end{tabular}
}
\end{table*}

\subsection{Transfer Across Generative Backbones}
\label{app:backbone_transfer}

We evaluate whether the benefit of pixel grounding extends beyond the FLUX backbone by implementing it on an SD2.1-based SR model. Table~\ref{tab:backbone_transfer_appendix} compares each backbone's latent-only SR baseline with its pixel-grounded counterpart on DIV2K-Val. On SD2.1, pixel grounding improves PSNR by 2.40 dB and SSIM by 0.0605, while reducing LPIPS from 0.2591 to 0.2316. The corresponding FLUX improvements are 2.21 dB, 0.0115, and 0.0327, respectively. These within-backbone comparisons support the contribution of pixel grounding beyond the choice of pretrained generative backbone.

\begin{table}[htbp]
\centering
\caption{Pixel grounding improves both distortion and perceptual metrics on DIV2K-Val with SD2.1 and FLUX. Bold denotes the better result within each backbone.}
\label{tab:backbone_transfer_appendix}
\small
\setlength{\tabcolsep}{5pt}
\begin{tabular}{llrrr}
\toprule
Backbone & Method & PSNR$\uparrow$ & SSIM$\uparrow$ & LPIPS$\downarrow$ \\
\midrule
SD2.1 & Latent-only SR baseline & 21.33 & 0.7048 & 0.2591 \\
SD2.1 & Baseline + pixel grounding & \textbf{23.73} & \textbf{0.7653} & \textbf{0.2316} \\
\midrule
FLUX.1-dev & Latent-only SR baseline & 23.68 & 0.7592 & 0.2431 \\
FLUX.1-dev & PGSR & \textbf{25.89} & \textbf{0.7707} & \textbf{0.2104} \\
\bottomrule
\end{tabular}
\end{table}

The transfer changes both the generative backbone and its native VAE tokenizer. Both tokenizers use an $f8$ spatial downsampling ratio, so this experiment evaluates transfer across backbones and tokenizers rather than across compression ratios. The magnitude of the gain varies between the two systems; the results do not imply equal effectiveness for arbitrary architectures.

\subsection{Training and Inference Cost}
\label{app:runtime_memory}

Table~\ref{tab:runtime_memory_appendix} compares PGSR with SinSR, OSEDiff, StableSR, and LucidFlux~\citep{fei2025lucidflux}. Inference timings use the same GPU, batch size 1, and a $448\times448\rightarrow1792\times1792$ SR task. We synchronize CUDA, exclude model loading, use one warm-up image, and average the next ten images. SinSR, OSEDiff, and StableSR use FP16; LucidFlux and PGSR use BF16, following the evaluated implementations. LucidFlux uses its official 20-step configuration and serves as a FLUX-based universal-restoration reference.

\begin{table}[htbp]
\centering
\caption{Measured inference latency and training cost. Time/step is the amortized time per image divided by the sampling-step count. Memory is peak training memory per GPU. LucidFlux is a universal-restoration model; the other methods are evaluated as SR systems.}
\label{tab:runtime_memory_appendix}
\setlength{\tabcolsep}{3.5pt}
\small
\resizebox{\textwidth}{!}{%
\begin{tabular}{llrrrrr}
\toprule
Method & Backbone & Steps & Time/step & Time/image & Train time/iter. & Peak train mem./GPU \\
\midrule
SinSR & ResShift & 1 & 2.14 s & 2.14 s & 0.67 s & 5.65 GiB \\
OSEDiff & SD2.1 & 1 & 9.79 s & 9.79 s & 1.11 s & 23.65 GiB \\
StableSR & SD2.1 & 20 & 4.58 s & 91.66 s & 1.27 s & 14.02 GiB \\
LucidFlux & FLUX.1-dev & 20 & 1.23 s & 24.61 s & 4.26 s & 35.54 GiB \\
PGSR & FLUX.1-dev & 20 & 3.09 s & 61.81 s & 3.59 s & 35.73 GiB \\
PGSR + sparse attention & FLUX.1-dev & 20 & 1.95 s & 38.95 s & 1.89 s & 39.14 GiB \\
\bottomrule
\end{tabular}%
}
\end{table}

PGSR remains slower per image than the one-step methods. Sparse attention reduces its inference latency from 61.81 s to 38.95 s (37.0\%) and training time per iteration from 3.59 s to 1.89 s, but increases peak training memory from 35.73 GiB to 39.14 GiB. Thus, the measured speed improvement does not imply a reduction in memory use. The training figures include the cost of decoding predictions for image-space supervision: although the VAE weights are frozen, back-propagation through the decoder still requires computation and activation memory. Per-iteration measurements describe the evaluated training configurations and do not establish equal total training budgets across methods.

Condition-side grounding and final decoder modulation are compatible in principle with fewer sampling steps. However, a distilled one-step PGSR model has not been evaluated, and its latency or reconstruction quality cannot be inferred from the amortized time/step alone.

\subsection{Incremental Computational Cost of Pixel Grounding}
\label{app:grounding_flops}

We profile the latent-only FLUX SR pipeline and its pixel-grounded variants under a $448\times448\rightarrow1792\times1792$ protocol with 20 sampling steps and 16 tiles of size $512\times512$. One multiply--add counts as two FLOPs. Table~\ref{tab:grounding_flops_appendix} counts the LR pixel extractor and condition fusion once per tile, with their outputs reused across sampling steps.

\begin{table}[htbp]
\centering
\caption{Inference FLOPs under a shared tiled-evaluation protocol. Relative changes use the latent-only FLUX SR pipeline as the baseline. These operation counts quantify incremental grounding cost rather than wall-clock acceleration.}
\label{tab:grounding_flops_appendix}
\small
\setlength{\tabcolsep}{8pt}
\begin{tabular}{lrr}
\toprule
Variant & TFLOPs/image & Relative to baseline \\
\midrule
Latent-only FLUX SR baseline & 7587.455 & $0\%$ \\
Condition-side only & 7587.559 & $+0.0014\%$ \\
Decoder-side only & 7588.899 & $+0.0190\%$ \\
Full PGSR & 7588.899 & $+0.0190\%$ \\
Full PGSR + sparse attention & 7579.106 & $-0.1100\%$ \\
\bottomrule
\end{tabular}
\end{table}

Full pixel grounding adds 1.444 TFLOPs per image, or approximately 0.019\% of the baseline cost. Decoder-side-only and full PGSR have identical rounded counts because they share the pixel extractor and the additional fusion cost is negligible at this scale. For comparison, an evaluator that recomputes the extractor and fusion at every denoising step requires 7590.862 TFLOPs/image ($+0.0449\%$). The small incremental cost under both accounting choices indicates that repeated FLUX sampling dominates the operation count. The sparse-attention FLOPs reduction is much smaller than its measured latency reduction in Table~\ref{tab:runtime_memory_appendix}; FLOPs and runtime therefore describe different aspects of efficiency.

\subsection{Decoder Feature Sources and Modulation}
\label{app:decoder_design}

We study the Pixel Extractor + DFM design through a controlled decoder-only reconstruction test. All variants receive the same clean-HR-derived latent from the FLUX VAE, removing differences in the flow trajectory. The VAE encoder and decoder remain frozen. We compare the native decoder, a CFW-style fusion reference adapted from StableSR, DFM driven by intermediate VAE-encoder features, and DFM driven by PGSR's independent pixel extractor. CFW-style is a reference modulation operator, rather than a component proposed by PGSR or a reproduction of StableSR's complete SR pipeline.

\begin{table}[htbp]
\centering
\caption{Decoder-only reconstruction comparison. Parameters count only trainable modules; the frozen VAE is excluded. The final row includes both the pixel extractor and DFM. Bold denotes the best reconstruction metric.}
\label{tab:decoder_design_appendix}
\small
\setlength{\tabcolsep}{4pt}
\resizebox{\textwidth}{!}{%
\begin{tabular}{llrrrr}
\toprule
Variant & Auxiliary feature & Trainable params. & Rec. PSNR$\uparrow$ & LPIPS$\downarrow$ & DISTS$\downarrow$ \\
\midrule
Frozen FLUX VAE & None & 0.00 M & 33.30 & 0.0190 & 0.0056 \\
Frozen VAE + CFW-style & VAE encoder taps & 105.82 M & 35.66 & 0.0222 & 0.0116 \\
Frozen VAE + DFM & VAE encoder taps & 4.92 M & 36.38 & 0.0157 & 0.0089 \\
Frozen VAE + Pixel Extractor + DFM & Pixel-extractor taps & 2.85 M & \textbf{38.09} & \textbf{0.0125} & \textbf{0.0051} \\
\bottomrule
\end{tabular}%
}
\end{table}

With identical VAE-encoder taps and decoder injection sites, replacing CFW-style fusion with DFM improves reconstruction PSNR from 35.66 dB to 36.38 dB and reduces LPIPS/DISTS from 0.0222/0.0116 to 0.0157/0.0089, using approximately $21.5\times$ fewer trainable parameters. The complete Pixel Extractor + DFM design further reaches 38.09 dB, 0.0125 LPIPS, and 0.0051 DISTS. This is the best reconstruction result among the tested variants and uses the fewest parameters among the trainable alternatives.

The 2.85 M count includes the pixel extractor (approximately 0.436 M) and DFM (2.412 M). The $32/64/256$-channel pixel taps require smaller adapters than the $128/256/512$-channel VAE taps (4.918 M). Thus, this comparison evaluates the combined feature-extraction and adapter design, not feature source at fixed width. The conclusions concern decoder reconstruction rather than end-to-end SR.

\section{Supplementary Analytical Discussion}
\label{app:theory}

This section provides additional analysis for the pixel-grounded components introduced in Sec.~3.
The goal is not to establish global optimality, but to clarify the inductive biases behind the proposed design: gated residual condition fusion can attenuate unreliable pixel corrections while keeping the LR latent as an anchor, multi-scale pixel taps reduce the bottleneck imposed by a single flat pixel representation, and zero-initialized DFM preserves the frozen decoder before image-space supervision becomes reliable.

\subsection{Gated Residual Condition Fusion}
\label{app:condition_fusion}

In Sec.~3, the condition-side fused latent is written as
\[
\tilde{z}_{lr}=z_{lr}+\lambda\,\sigma(g)\,W_p(\bar{p}_{lr}),
\]
where $z_{lr}$ is the VAE-encoded LR condition, $\bar{p}_{lr}$ is the pre-VAE pixel feature projected to the latent resolution, $W_p$ is a learned $1\times1$ projection, $g$ is a learnable scalar gate, and $\lambda$ controls the strength of the pixel evidence.

\paragraph{Proposition \thesection.1.}
Assume that the ideal fused condition is $z_{lr}+\delta$, and the projected pixel correction is a noisy estimate $W_p(\bar{p}_{lr})=\delta+\eta$ with $\mathbb{E}[\eta]=0$ and $\mathbb{E}\|\eta\|_2^2=\nu$.
Among residual fusions of the form $z_{lr}+\lambda\gamma W_p(\bar{p}_{lr})$ with $\gamma\in[0,1]$, the optimal gate is
\[
\gamma^\star
=
\mathrm{clip}_{[0,1]}
\left(
\frac{\|\delta\|_2^2}{\lambda(\|\delta\|_2^2+\nu)}
\right),
\]
and is no worse in expected squared error than any fixed additive choice when that fixed choice lies in the same interval.

\paragraph{Proof.}
For a given $\gamma$, the expected error of the fused condition is
\[
\mathcal{E}(\gamma)
=
\mathbb{E}
\left\|
z_{lr}+\lambda\gamma(\delta+\eta)-(z_{lr}+\delta)
\right\|_2^2
=
(1-\lambda\gamma)^2\|\delta\|_2^2+\lambda^2\gamma^2\nu .
\]
Minimizing this convex quadratic and projecting the minimizer onto $[0,1]$ gives $\gamma^\star$.
Since the gated family contains all fixed residual scales in $[0,1]$, its best attainable error cannot be larger than that of direct addition with any such fixed scale.
\hfill$\square$

\paragraph{Discussion.}
This proposition should be read as a bias argument rather than a full optimization guarantee.
The LR latent remains a stable residual anchor for the flow-matching trajectory, while the gate gives the model a simple mechanism to attenuate noisy pixel evidence under severe degradation.
Compared with fixed additive fusion, gated residual fusion keeps the benefits of immediate pixel participation but avoids forcing every pixel correction to enter the latent condition with the same strength.

\subsection{Approximation Advantage of Multi-Scale Pixel Taps}
\label{app:multiscale_taps}

PGSR uses a three-stage pixel extractor that produces matched taps
$\mathcal{S}_{lr}=\{s_1,s_2,s_3\}$ at $H/2$, $H/4$, and $H/8$ resolutions.
This design is motivated by the decoder-side use case.
The frozen latent decoder reconstructs an HR image through multiple upsampling blocks, and each block operates at a different spatial scale.
A flat extractor that outputs only one latent-resolution feature must predict corrections for all decoder stages from the same compressed representation.

Let $r_\ell$ denote the ideal pixel correction for decoder stage $\ell$.
A flat extractor first maps the upsampled LR image into a single low-resolution feature $q$ and then predicts each stage correction from $q$:
\[
q = B_3 x_{lr}^{\uparrow},
\qquad
\hat r_{\ell}^{\mathrm{flat}} = U_{\ell} A_{\ell} q,
\]
where $B_3$ is the low-resolution pixel encoder, $A_\ell$ is a stage-specific projection, and $U_\ell$ upsamples the feature when required.
The best achievable squared approximation error is
\[
\mathcal{E}_{\mathrm{flat}}
=
\min_{\{A_{\ell}\}}
\sum_{\ell}
\mathbb{E}
\left[
\left\|
r_{\ell} - U_{\ell} A_{\ell} B_3 x_{lr}^{\uparrow}
\right\|_2^2
\right].
\]
By contrast, a multi-scale extractor predicts each correction from a matched feature tap:
\[
s_{\ell}=B_{\ell}x_{lr}^{\uparrow},
\qquad
\hat r_{\ell}^{\mathrm{multi}} = A_{\ell}s_{\ell},
\]
with approximation error
\[
\mathcal{E}_{\mathrm{multi}}
=
\min_{\{A_{\ell},B_{\ell}\}}
\sum_{\ell}
\mathbb{E}
\left[
\left\|
r_{\ell} - A_{\ell} B_{\ell} x_{lr}^{\uparrow}
\right\|_2^2
\right].
\]

\paragraph{Proposition \thesection.2.}
The multi-scale formulation is at least as expressive as the flat formulation:
\[
\mathcal{E}_{\mathrm{multi}} \le \mathcal{E}_{\mathrm{flat}}.
\]

\paragraph{Proof.}
The multi-scale family can emulate any flat solution by choosing each $B_\ell$ to factor through the same low-resolution representation $B_3 x_{lr}^{\uparrow}$ and absorbing the upsampling/projection into $A_\ell$.
Therefore, the hypothesis class optimized by $\mathcal{E}_{\mathrm{flat}}$ is a subset of the hypothesis class optimized by $\mathcal{E}_{\mathrm{multi}}$.
Minimizing over a superset cannot yield a larger minimum error, which gives $\mathcal{E}_{\mathrm{multi}} \le \mathcal{E}_{\mathrm{flat}}$.
\hfill$\square$

\paragraph{Discussion.}
This result should be interpreted as an approximation-capacity argument rather than a guarantee of better test performance.
It supports the architectural choice of feeding DFM with scale-matched taps: high-frequency LR-observed cues may be attenuated by a latent-resolution bottleneck, and later projections cannot recover information that has already been discarded.
Providing $s_1$, $s_2$, and $s_3$ gives the decoder access to pixel evidence at the same scales where details are rendered.

\subsection{Identity Property of Decoder-Side DFM}
\label{app:dfm_identity}

For decoder stage $\ell$, the DFM adapter applies
\[
\tilde{f}_d^{\ell}
=
f_d^{\ell} + Z_{\ell}\!\left(\mathrm{SiLU}(A_{\ell}(s_{\ell}))\right),
\]
where $f_d^{\ell}$ is the frozen decoder activation, $s_{\ell}$ is the matched pixel feature tap, $A_\ell$ aligns channel width, and $Z_\ell$ is a zero-initialized output projection.

\paragraph{Proposition \thesection.3.}
If $Z_\ell$ is zero-initialized, then DFM preserves the frozen decoder at initialization:
\[
\tilde{f}_d^{\ell}=f_d^{\ell}
\quad
\mbox{for every decoder stage } \ell.
\]

\paragraph{Proof.}
Since $Z_\ell$ is zero-initialized,
\[
Z_{\ell}\!\left(\mathrm{SiLU}(A_{\ell}(s_{\ell}))\right)=0
\]
at initialization for any input feature $s_\ell$.
Therefore,
\[
\tilde{f}_d^{\ell}
=
f_d^{\ell}+0
=
f_d^{\ell}.
\]
\hfill$\square$

\paragraph{Discussion.}
The DFM-enhanced decoder starts exactly as the frozen VAE decoder and gradually learns pixel-grounded rendering only when image-space supervision provides a useful signal.
This is important because the decoder is a pretrained component: without an identity-preserving initialization, injected pixel modulation could perturb decoder activations before the adapters learn meaningful spatial corrections.





\end{document}